\documentclass[twocolumn]{article} %
\usepackage{arxiv}

\usepackage{times}  %
\usepackage{helvet}  %
\usepackage{courier}  %
\usepackage[hyphens]{url}  %
\usepackage{graphicx} %
\usepackage{natbib}  %
\usepackage{caption} %
\usepackage{algorithm}
\usepackage{algorithmic}

\usepackage{newfloat}
\usepackage{listings}

\DeclareCaptionStyle{ruled}{labelfont=normalfont,labelsep=colon,strut=off} %
\floatstyle{ruled}
\newfloat{listing}{tb}{lst}{}
\floatname{listing}{Listing}
\usepackage{amssymb}
\usepackage{amsmath}
\usepackage{amsthm}
\usepackage{refcount}
\usepackage{bm}
\usepackage{tabularx}
\usepackage{xcolor}
\usepackage{authblk}
\usepackage{url}

\newtheorem{theorem}{Theorem}[section] 
\newtheorem{corollary}{Corollary}[section]

\newcommand{\commentb}[1]{}
\newcommand{\commentsig}[1]{}
\newcommand{\commentsb}[1]{}
\newcommand{\commentsh}[1]{}
\newcommand{\commentsn}[1]{}
\newcommand{\commentpp}[1]{}
\newcommand{\adnote}[1]{\textcolor{blue}{}}
\newcommand{\commentp}[1]{}

\newcommand{\commentbold}[1]{}
\newcommand{\JS}[1]{}

\newcommand{\commentad}[1]{}
\newcommand{\commentunresolved}[1]{}

\newcommand{\ol}[1]{}

\usepackage{scalerel}
\usepackage{amsfonts}
\usepackage{mathtools}
\usepackage{calc}

\makeatletter
\newcommand{\mywidehat}[1]{%
  \ifmmode
    \ThisStyle{\mathrlap{\widehat{\phantom{\scalebox{0.82}{$A^*_r$}}}}{\scalebox{0.87}{$\SavedStyle#1$}}}%
  \else
    \textbf{#1} %
  \fi
}
\makeatother

\makeatletter
\newcommand{\myboldwidehat}[1]{%
  \ifmmode
        \ThisStyle{\mathrlap{\widehat{\phantom{\raisebox{0.25\LMex}{\scalebox{0.72}{$A^*_r$}}}}}{\scalebox{0.87}{$\SavedStyle#1$}}}
  \else
    \textbf{#1} %
  \fi
}
\makeatother

\newcommand{\adv}{A^*_r}
\newcommand{\approxa}{\mywidehat{A^*_r}}

\newcommand{\approxaraised}{\scalebox{0.5}{$\widehat{A^*_{\raisebox{0.2ex}{\scalebox{0.7}{$r$}}}}$}}
\newcommand{\arew}{r_{\adv}}
\newcommand{\approxarew}{r_{\approxaraised}}
\newcommand{\greedya}{greedy~\adv}
\newcommand{\greedyarew}{greedy~Q^*_{\raisebox{0.2ex}{\scalebox{0.8}{$\arew$}}}}
\newcommand{\greedyapproxa}{greedy~\approxa}

\newcommand{\greedyapproxarew}{greedy~Q^*_{\raisebox{0.2ex}{\scalebox{0.8}{$\approxarew$}}}}
\newcommand{\pistarsarew}{\Pi^*_{\raisebox{0.2ex}{\scalebox{0.7}{$\arew$}}}}
\newcommand{\pistarsapproxarew}{\Pi^*_{\raisebox{0.2ex}{\scalebox{0.7}{$\approxarew$}}}}

\newcommand{\partret}[2]{\Sigma_{\sigma_{#1}}{#2}}

\newcommand{\regretd}[2]{regret_{\text{d}}(\sigma_{#1}|#2)}

\newcommand{\valstart}[2]{V_{#2}^{*}(s^{\sigma_{#1}}_0)}
\newcommand{\valend}[2]{V_{#2}^{*}(s^{\sigma_{#1}}_{|\sigma_{#1}|})}

\DeclareTextCommand{\k}{OT1}[1]{{\oalign{#1\crcr\hidewidth\lower1ex\hbox{'}\hidewidth}}}

\title{Learning Optimal Advantage from Preferences and Mistaking it for Reward}

\renewcommand\Authfont{\bfseries}
\author[1,2]{{\hspace{1mm} W. Bradley Knox\thanks{\texttt{Correspondence to: bradknox@cs.utexas.edu}}}%
}
\author[1]{{\hspace{1mm} Stephane Hatgis-Kessell}}%
\author[2]{{\hspace{1mm} Sigurdur Orn Adalgeirsson}}%
\author[3]{{\hspace{1mm} Serena Booth}}%
\author[4]{{\hspace{1mm} Anca Dragan}}%
\author[1,5]{{\hspace{1mm} Peter Stone}}%
\author[6]{{\hspace{1mm} Scott Niekum}}%

\affil[1]{University of Texas at Austin }
\affil[2]{Google Research}
\affil[3]{MIT CSAIL}
\affil[4]{UC Berkeley}
\affil[5]{Sony AI}
\affil[6]{University of Massachusetts Amherst}

\begin{document}

\let\cite\citep
\let\shortcite\citeyearpar
\setcitestyle{aysep={}}
\setlength\bibhang{0pt}

\date{}
\maketitle

\begin{abstract}
We consider algorithms for learning reward functions from human preferences over pairs of trajectory segments, as used in reinforcement learning from human feedback (RLHF).
Most recent work assumes that human preferences are generated based only upon the reward accrued within those segments, or their \textit{partial return}.
Recent work casts doubt on the validity of this assumption, proposing an alternative preference model based upon \textit{regret}. We investigate the consequences of assuming preferences are based upon partial return when they actually arise from regret. We argue that the learned function is an approximation of the optimal advantage function, $\approxa$, \textit{not} a reward function. 
We find that if a specific pitfall is addressed, this incorrect assumption is not particularly harmful, resulting in a highly shaped reward function. 
Nonetheless, this incorrect usage of $\approxa$ is less desirable than the appropriate and simpler approach of greedy maximization of $\approxa$. 
From the perspective of the regret preference model, we also provide a clearer interpretation of fine tuning contemporary large language models with RLHF.
This paper overall provides insight regarding why learning under the partial return preference model tends to work so well in practice, despite it conforming poorly to how humans give preferences.
\end{abstract}

\newpage
~
\vspace{76mm}
\section{Introduction}
\label{sec:intro}

\begin{figure}[ht]%
  \vspace{0mm}
  \centering
  \includegraphics[width=0.75\columnwidth]{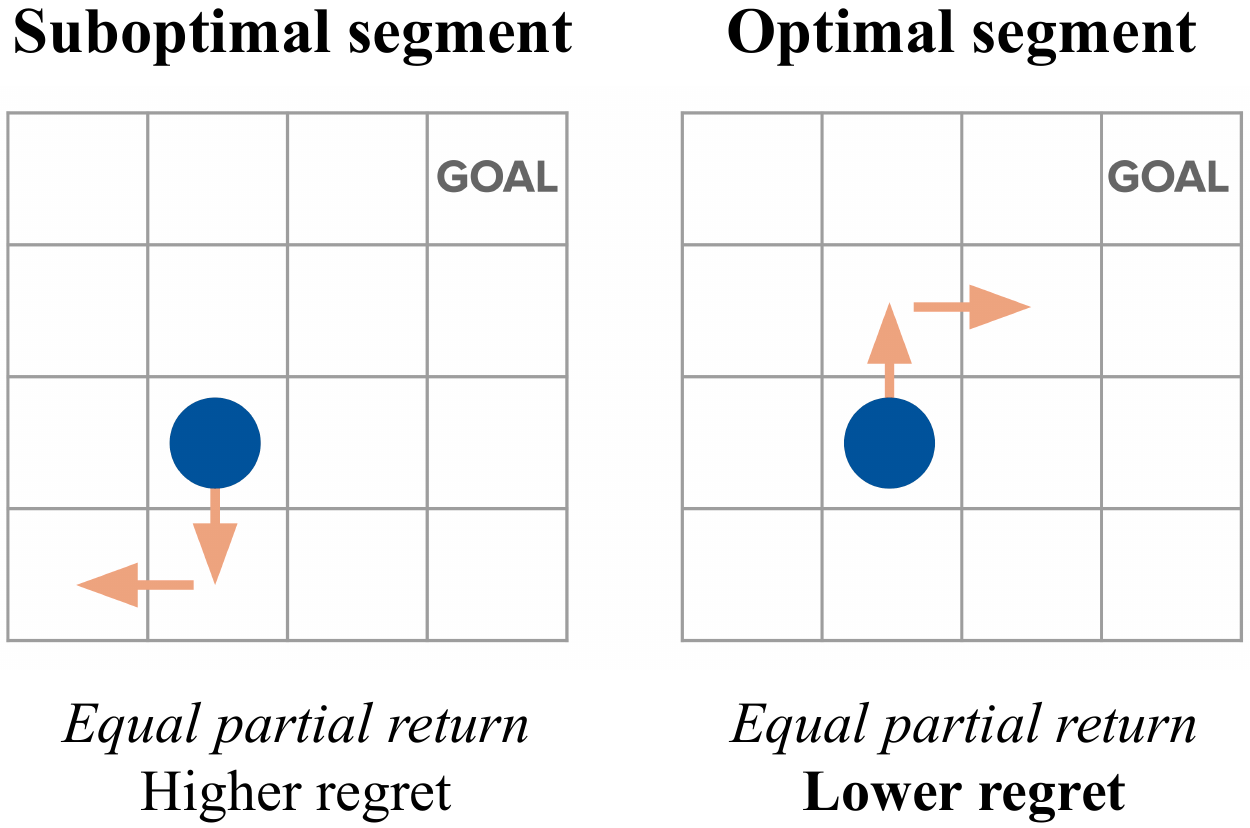}
  \vspace{0mm}
  \caption{\footnotesize Two segments in an undiscounted task with $-1$ reward each time step. The partial return of both segments with respect to the true reward function is $-2$. The regret of the left segment is $4$. The right segment is optimal and therefore has a regret of $0$. The regret preference model is more likely to prefer the right segment---as we suspect our human readers are---whereas the partial return preference model is equally likely to prefer each segment.}
  \label{fig:modelintuition}
  \vspace{0mm}
\end{figure}

When learning from human preferences (in RLHF), the dominant model assumes that human preferences are determined only by each segment's accumulated reward, or \textbf{partial return}. \citet{knox2022models} argued that the partial return preference model has fundamental flaws that are removed or ameliorated by instead assuming that human preferences are determined by the \textbf{optimal advantage} of each segment, which is a measure of deviation from optimal decision-making and is equivalent to the negated \textbf{regret}. This past work argues for the superiority of the regret preference model (1) by intuition, regarding how humans are likely to give preferences (e.g., see Fig.~\ref{fig:modelintuition}); (2) by theory, showing that regret-based preferences have a desirable identifiability property that preferences from partial return lack; (3) by descriptive analysis, showing that the likelihood of a human preferences dataset is higher under the regret preference model than under the partial return preference model; and (4) by empirical analysis, showing that with both human and synthetic preferences, the regret model requires fewer preference labels. Section~\ref{sec:prefmodels} of this paper provides details on the general problem setting and on these two models.

In this paper, we explore the consequences of using algorithms that are designed with the assumption that preferences are determined by partial return when these preferences are instead determined by regret. We show in Section~\ref{sec:oaf-as-rew} that these algorithms learn an approximation of the optimal \emph{advantage function}, $\adv$, not of the reward function, as presumed in many prior works. 
We then study the implications of this mistaken interpretation. When interpreted as reward, the \textit{exact} optimal advantage is highly shaped and preserves the set of optimal policies, which enables partial-return-based algorithms to perform well. 
However, the learned \textit{approximation} of the optimal advantage function, $\approxa$, will have errors. 
We characterize when and how such errors will affect the set of optimal policies with respect to this mistaken reward, and we uncover a method for reducing a harmful type of error.
We conclude that this incorrect usage of $\approxa$ still permits decent performance under certain conditions, though it is less desirable than the appropriate and simpler approach of greedy maximization of $\approxa$.

\begin{figure}[t]
   \vspace{0mm}
    \centering
    \includegraphics[width=1.0\columnwidth]{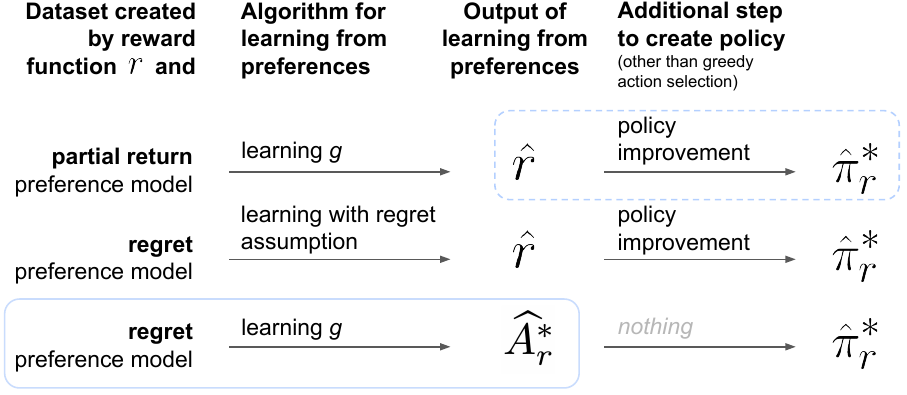}
    \vspace{0mm}
    \caption{\footnotesize Three algorithms that are justified by their assumed preference model. The top algorithm was popularized by \citet{christiano2017deep} and the middle algorithm was proposed by \citet{knox2022models}. The third algorithm is described in Section~\ref{sec:a-star_as_reward}. The reward function $\hat{r}$, optimal advantage function $\approxa$, and optimal policy $\hat{\pi}^*_r$ are approximations of the true versions of these functions. 
    The function $g$ is defined generally in Equation~\ref{eq:logisticonsummation} to allow it to represent including $\adv$ or $r$. This paper focuses on what occurs when the solid box represents the actual algorithm for learning $g$ but the partial return preference model is assumed, causing $\approxa$ to be used as if it is the reward in the dashed box.%
    }
    \label{fig:3algs}
    \vspace{0mm}
\end{figure}

We then show in Section~\ref{sec:reframing} that recent algorithms used to fine-tune state-of-the-art language models ChatGPT~\cite{openai2022chatgpt}, Sparrow~\cite{glaese2022improving}, and others~\cite{ziegler2019fine,ouyang2022training,bai2022training, touvron2023llama} can be viewed as an instance of learning an optimal advantage function and inadvertently treating it as one.
In multi-turn (i.e., sequential) settings such as that of ChatGPT, Sparrow, and research by~\citet{bai2022training},
this alternative framing allows the removal of a problematic assumption of these algorithms: that a reward function learned for a sequential task is instead used in a bandit setting, effectively setting the discount factor $\gamma$ to $0$.

\section{Preliminaries: Preference models for learning reward functions}
\label{sec:prefmodels}

A Markov decision process (MDP) is specified by a tuple ($S$, $A$, $T$, $\gamma$, $D_0$, $r$).  $S$ and $A$ are the sets of possible states and actions, respectively. $T: S \times A \rightarrow  p(\cdot | s,a)$ is a transition function;
$\gamma$ is the discount factor; and $D_0$ is the distribution of start states. Unless stated otherwise, we assume tasks are undiscounted ($\gamma=1$) and have terminal states, after which only $0$ reward can be received. $r$ is a reward function, $r: S \times A \times S \rightarrow \mathbb{R}$, where $r_t$ is a function of $s_t$, $a_t$, and $s_{t+1}$ at time $t$. An MDP$\setminus r$ is an MDP without a reward function. 

Throughout this paper, $r$ refers to the ground-truth reward function for some MDP; $\hat{r}$ refers to a learned approximation of $r$; and $\tilde{r}$ refers to any reward function (including $r$ or $\hat{r}$).
A policy (${\pi}: S \times A \rightarrow [0,1]$) specifies the probability of an action given a state. 
$Q^\pi_{\tilde{r}}$ and $V^\pi_{\tilde{r}}$ refer respectively to the state-action value function and state value function for a policy, $\pi$, under $\tilde{r}$, and are defined as follows.
\begin{equation*}
\begin{split}
V^\pi_{\tilde{r}}(s) \triangleq \mathbb{E}_{\pi}[\sum_{t=0}^{\infty} \tilde{r}(s_t, a_t, s_{t+1}) | s_0 = s]\\
Q^\pi_{\tilde{r}}(s,a) \triangleq \mathbb{E}_{\pi}[\tilde{r}(s,a,s') + V^\pi_{\tilde{r}}(s')]
\end{split}
\end{equation*}
An optimal policy $\pi^*_{\tilde{r}}$ is any policy where $V^{\pi^{*}_{\tilde{r}}}_{\tilde{r}}(s) \geq V^\pi_{\tilde{r}}(s)$ at every state $s$ for every policy $\pi$. We write shorthand for $Q^{\pi^*_{\tilde{r}}}_{\tilde{r}}$ and $V^{\pi^*_{\tilde{r}}}_{\tilde{r}}$ as $Q^*_{\tilde{r}}$ and $V^*_{\tilde{r}}$, respectively. The optimal advantage function is defined as $A^*_{\tilde{r}}(s,a) \triangleq Q^*_{\tilde{r}}(s,a) - V^*_{\tilde{r}}(s)$; this measures how much an action reduces expected return relative to following an optimal policy.

Throughout this paper, when the preferences are not human-generated, the ground-truth reward function $r$ is used to algorithmically generate preferences. $r$ is hidden during reward learning and is used to evaluate the performance of optimal policies under a learned $\hat{r}$.

\subsection{Reward learning from pairwise preferences}
\label{sec:prefdefs}

A reward function is commonly learned by minimizing the cross-entropy loss---i.e., maximizing the likelihood---of observed human preference labels~\cite{christiano2017deep,ibarz2018reward,wang2022skill,biyik2021learning,sadigh2017active,lee2021pebble,lee2021b,ziegler2019fine,ouyang2022training,bai2022training,glaese2022improving,openai2022chatgpt,touvron2023llama}. %

\textbf{Segments~~~} 
Let $\sigma$ denote a segment starting at state $s^{\sigma}_{ 0}$. Its length $|\sigma|$ is the number of transitions within the segment. A segment includes $|\sigma| + 1$ states and $|\sigma|$ actions: 
$(s^{\sigma}_{0},a^{\sigma}_{0}, s^{\sigma}_{1},a^{\sigma}_{1}, ..., s^{\sigma}_{{|\sigma|}})$.
In this problem setting, segments lack any reward information. 
As shorthand, we define $\sigma_t \triangleq (s^{\sigma}_{t}, a^{\sigma}_{t}, s^{\sigma}_{t+1})$. 
A segment $\sigma$ is \textbf{optimal} with respect to $\tilde{r}$ if, for every $i \in \{1, ..., |\sigma|\text{-}1\}$, $A^*_{\tilde{r}}(s^{\sigma}_{i},a^{\sigma}_{i}) = 0$. A segment that is not optimal is \textbf{suboptimal}. 
Given some $\tilde{r}$ and a segment $\sigma$, where
$\tilde{r}_t^{\sigma} \triangleq \tilde{r}(s^{\sigma}_{t}, a^{\sigma}_{t}, s^{\sigma}_{t+1})$, the undiscounted \textbf{partial return} of a segment $\sigma$ is $\sum_{t=0}^{|\sigma|-1} \tilde{r}_t^{\sigma}$, which we denote in shorthand as $\Sigma_{\sigma}\tilde{r}$.

\textbf{Preference datasets~~~}
Each preference over a pair of segments creates a sample $(\sigma_1, \sigma_2, \mu)$ in a preference dataset $D_{\succ}$. Vector $\mu = \langle \mu_1,\mu_2 \rangle$ represents the preference; specifically,
if $\sigma_1$ is preferred over $\sigma_2$, denoted $\sigma_1 \succ \sigma_2$, $\mu =\langle 1,0 \rangle$. $\mu$ is $\langle 0,1 \rangle$ if $\sigma_1 \prec \sigma_2$ and is $\langle0.5,0.5\rangle$ for $\sigma_1 \sim \sigma_2$ (no preference). %
For a sample $(\sigma_1, \sigma_2, \mu)$, we assume that the two segments have equal lengths (i.e., $|\sigma_1|=|\sigma_2|$).

\textbf{Loss function~~~}
When learning a reward function from a preference dataset, $D_{\succ}$, preference labels are typically assumed to be generated by a preference model $P$ based on an unobservable \textit{ground-truth} reward function $r$.%
We learn $\hat{r}$, an approximation of $r$, by minimizing cross-entropy loss: 
\begin{equation}
\label{eq:loss}
\begin{gathered}
loss(\hat{r},D_{\succ}) = \\
- \hspace{-7mm}\sum_{(\sigma_1, \sigma_2, \mu) \in D_{\succ}} \hspace{-5mm} \mu_1 \log {P}(\sigma_1 \succ \sigma_2 | \hat{r}) + \mu_2 \log {P}(\sigma_1 \prec \sigma_2 | \hat{r})
\end{gathered}
\end{equation}
If $\sigma_1 \succ \sigma_2$, the sample's likelihood is ${P}(\sigma_1 \succ \sigma_2 | \hat{r})$ and its loss is therefore $-log {P}(\sigma_1 \succ \sigma_2 | \hat{r})$. If $\sigma_1 \prec \sigma_2$, its likelihood is $1 - {P}(\sigma_1 \succ \sigma_2 | \hat{r})$.
This loss is under-specified until the preference model ${P}(\sigma_1 \succ \sigma_2 | \hat{r}$) is defined.  Algorithms in this paper for learning approximations of $r$ or $A^*_r$ from preferences can be summarized simply as ``minimize Equation~\ref{eq:loss}''.

\textbf{Preference models~~~}
A preference model determines the probability of one trajectory segment being preferred over another, $P(\sigma_1 \succ \sigma_2 | \tilde{r})$. Preference models can be used to model preferences provided by humans or other systems, or to generate synthetic preferences.

\subsection{Preference models: partial return and regret}
\label{sec:twoprefmodels}

\textbf{Partial return~~~}
The dominant preference model (e.g., \citet{christiano2017deep}) assumes human preferences are generated by a Boltzmann distribution over the two segments' partial returns, expressed here as a logistic function:\footnote{Unless otherwise stated, we ignore the temperature because scaling reward has the same effect as changing the temperature.}%
 \vspace{0mm}
\begin{equation}
\label{eq:partialreturn}
    P_{\Sigma_r}(\sigma_1 \succ \sigma_2 | \tilde{r}) = 
    logistic\Big(\partret{1}{\tilde{r}}-\partret{2}{\tilde{r}}\Big).%
\end{equation}
\textbf{Regret~~~} \citet{knox2022models} introduced an alternative human preference model. This regret-based model assumes that preferences are based on segments' deviations from optimal decision-making: the regret of each transition in a segment. We first focus on segments with deterministic transitions. For a single transition $(s_t,a_t,s_{t+1})$, 
$\regretd{t}{\tilde{r}} \triangleq V^*_{\tilde{r}}(s^{\sigma}_{t}) - [{\tilde{r}}_t + V^*_{\tilde{r}}(s^{\sigma}_{t+1})]$. For a full segment,
 \vspace{0mm}
\begin{equation}
\begin{split}
\label{eq:segmentregretdet}
regret_d(\sigma | \tilde{r}) &\triangleq \sum_{t=0}^{|\sigma|-1} regret_d(\sigma_t | \tilde{r}) \\ &= 
     \valstart{}{\tilde{r}} - (\partret{}{\tilde{r}} + \valend{}{\tilde{r}}),
\end{split}
\end{equation} 
with the right-hand expression arising from cancelling out intermediate state values. Therefore, deterministic regret measures how much the segment reduces expected return from $\valstart{}{{\tilde{r}}}$. An optimal segment $\sigma^*$ always has 0 regret, and a suboptimal segment  $\sigma^{\neg *}$ always has positive regret.

Stochastic state transitions, however, can result in  $regret_d(\sigma^*|\hat{r})>regret_d(\sigma^{\neg *}|\tilde{r})$, losing the property above. To retain it, we note that the effect on expected return of transition stochasticity from a transition $(s_t, a_t, s_{t+1})$ is $[{\tilde{r}}_t + V^*_{\tilde{r}}(s_{t+1})] - Q^*_{\tilde{r}}(s_t,a_t)$ and add this expression once per transition to get $regret(\sigma)$, removing the subscript $d$ that refers to determinism. 
The regret for a single transition becomes
$regret(\sigma_t | \tilde{r}) =  
\bm{[}V^*_{\tilde{r}}(s^{\sigma}_{t}) - [{\tilde{r}}_t + V^*_{\tilde{r}}(s^{\sigma}_{t+1})]\bm{]}
+ \bm{[}[{\tilde{r}}_t + V^*_{\tilde{r}}(s^{\sigma}_{t+1})] 
- Q^*_{\tilde{r}}(s^{\sigma}_{t},a^{\sigma}_{t})\bm{]} 
=
V^*_{\tilde{r}}(s^{\sigma}_{t}) - Q^*_{\tilde{r}}(s^{\sigma}_{t},a^{\sigma}_{t}) = -A^*_{\tilde{r}}(s^{\sigma}_{t},a^{\sigma}_{t})$.
Regret for a full segment is:
 \vspace{0mm}
\begin{equation}
\begin{split}
\label{eq:segmentregret}
    regret(\sigma | \tilde{r}) &= \sum_{t=0}^{|\sigma|-1} regret(\sigma_t | \tilde{r}) 
    \\&= \sum_{t=0}^{|\sigma|-1} \Big[V^*_{\tilde{r}}(s^{\sigma}_{t}) - 
    Q^*_{\tilde{r}}(s^{\sigma}_{t},a^{\sigma}_{t})\Big] 
    \\&= \sum_{t=0}^{|\sigma|-1} -A^*_{\tilde{r}}(s^{\sigma}_{t},a^{\sigma}_{t}).
\end{split}
\end{equation} 
The regret preference model is the Boltzmann distribution over the sum of optimal advantages, or the \textit{negated} regret:
\vspace{0mm}
\begin{equation}
\begin{split}
\label{eq:regretprefmodel}
P_{regret}&(\sigma_1 \succ \sigma_2 | \tilde{r}) \\
&\triangleq logistic\Big( 
\sum_{t=0}^{|\sigma_1|-1} A^*_{\tilde{r}}(\sigma_{1,t}) -
\sum_{t=0}^{|\sigma_2|-1} A^*_{\tilde{r}}(\sigma_{2,t})
\Big) \\
&= logistic\Big(regret(\sigma_2 | \tilde{r}) - regret(\sigma_1 | \tilde{r}) \Big).
\end{split}
\end{equation} 
(Notationally, $A^*_{\tilde{r}}(\sigma_t) = A^*_{\tilde{r}}(s^{\sigma}_t,a^{\sigma}_t)$.)
Lastly, if two segments have deterministic transitions, end in terminal states, and have the same starting state, this regret model reduces to the partial return model: $P_{regret}(\cdot|\tilde{r}) = P_{\Sigma_r}(\cdot|\tilde{r})$.%

Intuitively, the partial return preference model always assumes preferences are based upon outcomes while the regret model is able to account for preferences based upon outcomes (Eq. \ref{eq:segmentregretdet}) and preferences over decisions (Eq. \ref{eq:segmentregret}).

\citet{knox2022models} showed the regret both has desirable theoretical properties (i.e., it is identifiable where partial return is not) and is a better model of true human preferences.
Since regret better models true human preferences, and since many recent works use true human preferences but assume them to be generated according to partial return, we ask: what are the consequences of misinterpreting the optimal advantage function as reward?

\section{Learning optimal advantage from preferences and using it as reward}
\label{sec:oaf-as-rew}

We ask: what is actually learned when preferences are assumed to arise from partial return but actually come from regret (Equation~\ref{eq:partialreturn}), and what implications does that have?

Our results can be reproduced via our code repository, at \url{github.com/Stephanehk/Learning-OA-From-Prefs}.

\subsection{Learning the optimal advantage function}
To start, let us unify the two preference models from Section~\ref{sec:twoprefmodels} into a single general preference model. 

\vspace{0mm}
\begin{equation}
\label{eq:logisticonsummation}
    \hspace{-0.5mm}P_g(\sigma_1 \succ \sigma_2|\tilde{r}) \triangleq logistic\Big(\hspace{-0.4mm} \sum_{t=0}^{|\sigma_1|-1} g(\sigma_{1,t}) - \sum_{t=0}^{|\sigma_2|-1} g(\sigma_{2,t}) \Big) %
\end{equation}
\hspace{-0.2mm}In the above unification, the segment statistic in the preference model is expressed as a sum of some function $g$ over each transition in the segment: $\sum_{t=0}^{|\sigma|-1} g(\sigma_t) = \sum_{t=0}^{|\sigma|-1} g(s^{\sigma}_{t}, a^{\sigma}_{t}, s^{\sigma}_{t+1})$. 
When preferences are generated according to partial return, $g(\sigma_t) = \tilde{r}(s^{\sigma}_{t}, a^{\sigma}_{t}, s^{\sigma}_{t+1})$, and the reward function $\tilde{r}$ is learned via Equation~\ref{eq:loss}.

When preferences are instead generated according to regret, $g(\sigma_t) = 
A^*_{\tilde{r}}(\sigma_t) = A^*_{\tilde{r}}(s^{\sigma}_{t}, a^{\sigma}_{t})$ and the parameters of this optimal advantage function can be learned directly, 
also via Equation~\ref{eq:loss}.  $\approxa$ can be learned and then acted upon greedily, via $argmax_a \approxa(s,a)$, an algorithm we call $\bm{{\greedyapproxa}}$ (bottom algorithm of Fig.~\ref{fig:3algs}). Notably, this algorithm does not require the additional step of policy improvement and instead uses $\approxa$ directly. 
No reward function is explicitly represented or learned, though we still assume that preferences were generated by regret under a hidden reward function $r$.

The remainder of this section considers first the consequences of using the error-free $\adv$ as a reward function: $\bm{r_{\adv}=\adv}$. We call this mistaken approach $\bm{\greedyarew}$. We then consider the consequences of using the approximation $\approxa$ as a reward function, $\approxarew=\approxa$, which we refer to as $\bm{\greedyapproxarew}$. 
The following investigation is an attempt to answer \textit{why learning while assuming the partial return preference model tends to work so well in practice, despite its poor fit as a descriptive model of human preference.}

\subsection{Using $\bm{{\adv}}$  as a reward function}
\label{sec:a-star_as_reward}

Under the assumption of regret-based preferences, learning a reward function with the partial return preference model effectively uses an approximation of $\adv$ as a reward function, $\hat{r} = \approxa$. Let us first assume perfect inference of $\adv$ (i.e., that $\approxa=\adv$), and consider the consequences. We will refer to the \textit{non-approximate} versions of $\greedyapproxa$ and $\approxarew$ as  $\bm{\greedya}$ and $\bm{\arew}$.

\paragraph{Optimal policies are preserved.}
Using $\adv$ as a reward function preserves the set of optimal policies. 
To prove this statement, we first prove a more general theorem. 

For $\tilde{r}$, an arbitrary reward function, $max_a {A}^*_{\tilde{r}}(\cdot,a)=0$ by definition. 
Let the set of optimal policies with respect to $\tilde{r}$ be denoted $\Pi^*_{\tilde{r}}$.

\begin{theorem} [Greedy action is optimal when the maximum reward in every state is 0.]
\label{thm:myopic_is_optimal}
~ \\  
$\Pi^*_{\tilde{r}} = \{\pi : \forall s, \forall a ~ [\pi(a|s) > 0 \Leftrightarrow a \in \text{argmax}_a \tilde{r}(s,a)] \}$ if $\text{max}_a \tilde{r}(\cdot,a)=0$.
\end{theorem}

\vspace{1mm}\noindent Theorem~\ref{thm:myopic_is_optimal} is proven in Appendix~\ref{app:theorem_proof}. The sketch of the proof is that if the maximum reward in every state is 0, then the best possible return from every state is 0. Therefore, $V^*_{\tilde{r}}(\cdot)=0$, making $\forall (s,a) \in S \times A, Q^*_{\tilde{r}}(s,a) = \tilde{r}(s,a) + \gamma \mathbb{E}_{s'}[V^*_{\tilde{r}}(s)] = \tilde{r}(s,a)$.

We now return to our specific case, proven in Appendix~\ref{app:corollary_proof}.

\begin{corollary} [Policy invariance of $\arew$]
\label{thm:policy_invariance} ~\\
    Let $\arew \triangleq \adv$.
    If $\text{max}_a \adv(\cdot,a)=0$, 
    $\pistarsarew = \Pi^*_{r}$.
\end{corollary}

\paragraph{An underspecification issue is resolved.}
As we discuss in Section~\ref{sec:reframing}, when segment lengths are 1, the partial return preference model ignores the discount factor $\gamma$, making its choice arbitrary despite it often affecting the set of optimal policies. With $\arew$, however, the lack of $\gamma$ in Corollary~\ref{thm:policy_invariance} establishes $\gamma$ does not affect the set of optimal policies. To give intuition, we apply the intermediate result within the proof of Theorem~\ref{thm:myopic_is_optimal} that $V^*_{\tilde{r}}(\cdot)=0$ to the specific case of Corollary~\ref{thm:policy_invariance}, we see that $V^*_{\arew}(\cdot)=0$. Therefore, $Q^*_{\arew}(s,a) = \arew(s,a) + \gamma \mathbb{E}_{s'}[0]$, making $\gamma$ have no impact on $Q^*_{\arew}(s,a)$ and therefore on  on $\Pi^*_{r}$.

\paragraph{Reward is highly shaped.}
In \citet{ng1999piu}'s seminal research on potential-based reward shaping , they highlight $\phi(s) = V_{r}^*(s)$ as a particularly desirable potential function. 
Algebraic manipulation reveals that the MDP that results from this $\phi$ actually uses as a reward function $\arew \triangleq A^*_{r}$. See Appendix~\ref{app:shaped} for the derivation.
Ng et al. also note that that it causes $V_{\arew}^*(\cdot)=0$ and therefore results in ``a particularly easy value function to learn; ... all that would remain to be done would be to learn the non-zero Q-values.'' We characterize this approach as highly shaped because the information required to act optimally is in the agent's immediate reward.

\vspace{0mm}
\paragraph{Policy improvement wastes computation and environment sampling.}
When using $\adv$ as a reward function, no policy improvement is needed: setting
$\pi(s) = argmax_a [\adv(s,a)]$ provides an optimal policy.

\subsection{Using the learned $\bm{{\myboldwidehat{A^*_r}}}$ as a reward function}
\label{sec:approx-oaf-as-rew}

A caveat to the preceding analysis is that the algorithm does not necessarily learn $\adv$. Rather it learns its approximation, $\approxa$. 
We investigate the effects of the approximation error of $\approxa$. We find that this error only induces a difference in performance from that of $\greedyapproxa$ when $max_a \approxa(s,a) \neq 0$ in at least one state $s$, and the consequence of that error is dependent on the maximum partial return of all \textit{loops}---segments that start and end in the same state---within the MDP.

For the empirical results below, we build upon the experimental setting of \citet{knox2022models}, including both for learning and for randomly generating MDPs. Hyperparameters and other experimental settings are identical except where noted. All preferences are synthetically generated by the regret preference model.%

\vspace{-1mm}
\paragraph{If the maximum value of %
$\bm{{\myboldwidehat{A^*_r}}}$ in every state is 0, behavior is identical between $\bm{\greedyapproxarew}$ and $\bm{greedy~{\myboldwidehat{A^*_r}}}$.}

From Theorem~\ref{thm:myopic_is_optimal}, the following trivially holds for a learned approximation $\approxa$.
\begin{corollary} %
\label{thm:max0} Let $\approxarew \triangleq \approxa$.
    If $\text{max}_a \approxa(\cdot,a)=0$, then 
    $\pistarsapproxarew = \{\pi : \forall s, \forall a ~ [\pi(a|s) > 0 \Leftrightarrow a \in \text{argmax}_a \approxa(s,a)] \}$.
\end{corollary}

\noindent
Therefore, if $max_a \approxa(\cdot,a)=0$, then a policy from $\greedyapproxa$ is identical to an optimal policy for $\greedyapproxarew$, assuming ties are resolved identically. The actual policy from $\greedyapproxarew$ will also be identical unless limitations of the policy improvement algorithm cause it to not find a policy in $\pistarsapproxarew$ in this highly shaped setting with the reward function also in hand, not requiring experience to query. %
However, $max_a \approxa(\cdot,a)=0$ is not guaranteed for an approximation of $\adv$, which we consider later in this section.

We conduct an empirical test of the assertion above by adjusting $\approxa$ to have the property $max_a \approxa(\cdot,a)=0$ by shifting $\approxa$ by a state-dependent constant: for all $(s,a)$, 
$r_{\approxaraised{\scalebox{0.5}{{-}$shifted$}}}(s,a) \triangleq \approxa(s,a) - max_{a'} \approxa(s,a')$. 
Note that $argmax_a r_{\approxaraised\scalebox{0.5}{{-}$shifted$}}(s,a)%
= argmax_a \approxa(s,a)$.
In 90 small gridworld MDPs, we observe no difference between $\greedyapproxa$ and $\greedyapproxarew$ with $r_{\approxaraised\scalebox{0.5}{{-}$shifted$}}$ (see Figure~\ref{fig:max_a_0_confirmation}).  However, cost is generally incurred from suboptimal behavior and environment sampling \textit{while} a policy improvement algorithm learns this approximately optimal policy, unless the policy improvement algorithm uses the in-hand $r_{\approxaraised\scalebox{0.5}{{-}$shifted$}}$ without environment sampling and makes use of knowledge that the state value is 0 in every state, which together allow it to simply define optimal behavior as $argmax_a Q_{r_{\approxaraised\scalebox{0.5}{{-}$shifted$}}}(s,a) = argmax_a r_{\approxaraised\scalebox{0.5}{{-}$shifted$}}(s,a)%
= argmax_a \approxa(s,a)$, which is $\greedyapproxa$.

\paragraph{Including segments with transitions from absorbing state encourages $\bm{max_a {\myboldwidehat{A^*_r}}(\cdot,a)=0}$.}

If an algorithm designer is confident that the preferences in their preference dataset were generated via the regret preference model, then the technique above of manually shifting $\approxa$ may be justified and tenable, depending on upon the size of the action space. Yet with such confidence, acting to greedily maximize $\approxa$ is more straightforward and efficient. Further, an appeal that will emerge from our analysis is that \textit{algorithmically} assuming preferences arise from partial return can lead to good performance regardless of whether preferences actually reflect partial return or regret. The manual shift technique could change the set of optimal policies when preferences are generated by the partial return preference model. Therefore, we do not recommend applying the shift above in practice. Below we describe another method that, although imperfect, avoids explicitly embracing either preference model.

\begin{figure}[t]
    \vspace{0mm}
    \centering
    \includegraphics[width=1.0\columnwidth]{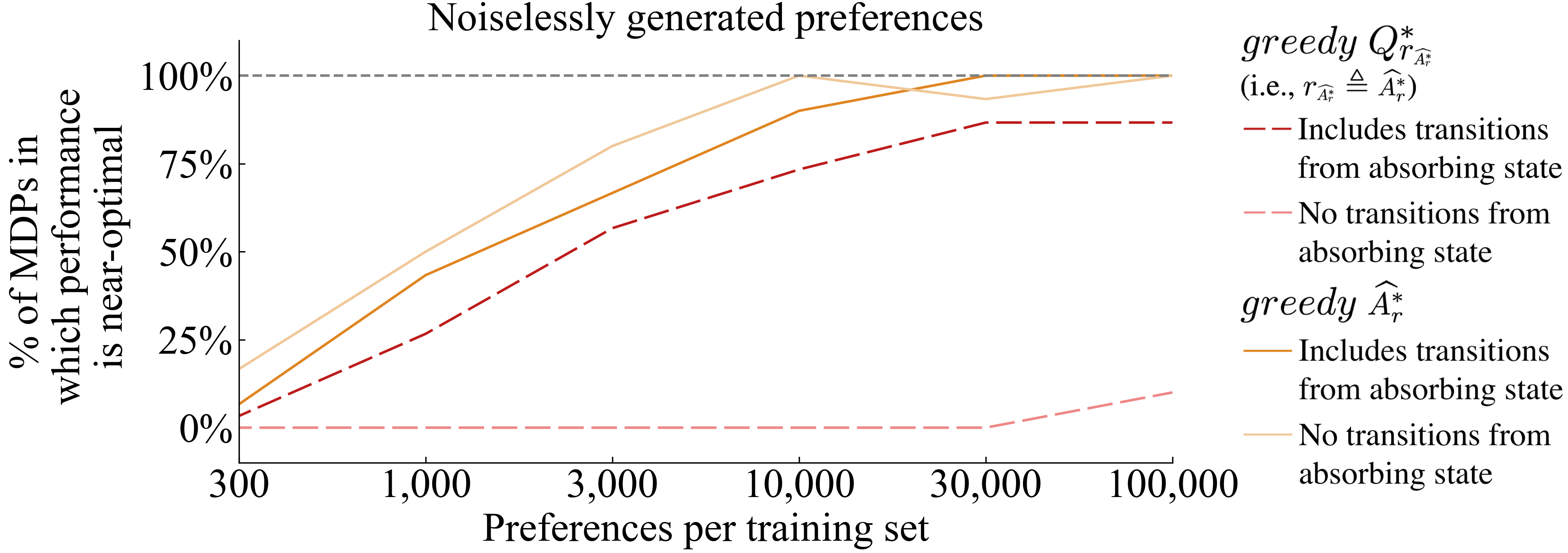}
    \\ 
    \vspace{0mm}
    \caption{\footnotesize Performance when noiselessly generated preference datasets do and do not include segments with transitions from absorbing state. Results are across 30 randomly generated gridworld MDPs with tabular representations of the $\approxa$, where segments of length 3 are chosen by uniformly randomly choosing a start state and 3 its actions.
    When transitions from absorbing states are not included, any segment that terminates before its final transition is rejected and then resampled. 
    For $\greedyapproxa$ (in red) Wilcoxon paired signed-rank tests reveal that including transitions from absorbing state results in significantly higher performance for all training set sizes but the smallest, 300, with $p < 0.0007$. No significant difference in performance is detected for $\greedyapproxarew$ with or without terminating transitions except at 30,000 preferences with a more modest $p = 0.04$.
    Appendix~\ref{app:performance_diff_inv} contains the plot for stochastically generated preferences (Figure~\ref{fig:absorbing_state_stochastic}), which contains similar results.
    }
    \label{fig:absorbing_state_noiseless}
    \vspace{-4mm}
\end{figure}

Adding a constant to $\approxa$ does not change the likelihood of a preferences dataset, making the \textit{learned} value of $max_{(s,a)} \approxa(s,a)$ arbitrary. Consequently, it also makes $max_a \approxa(\cdot,a)$ underspecified. If tasks have varying horizons, then different choices for this maximum value can determine different sets of optimal policies (e.g., by changing whether termination is desirable). One solution is to convert varying horizon tasks to continuing tasks by including infinite transitions from absorbing states to themselves after termination, where all such transitions receive $0$ reward. Note that this issue does not exist when acting directly from $\approxa$---i.e., $\pi(s) = argmax_a [\approxa(s,a)]$---for which adding a constant to the output of $\approxa$ does not change $\pi$.
Some past authors have acknowledged this insensitivity to a shift ~\citep{christiano2017deep,lee2021pebble,ouyang2022training,hejna2023inverse}, and the common practice of forcing all tasks to have a fixed horizon (e.g., as done by ~\citet[p. 14]{christiano2017deep} and \citet{gleave2022imitation}) may be partially attributable to the poor performance that results when using the partial return preference model in variable-horizon tasks without transitions from absorbing states.

\begin{figure}[t]
    \vspace{0mm}
    \centering
    \includegraphics[width=1\columnwidth]{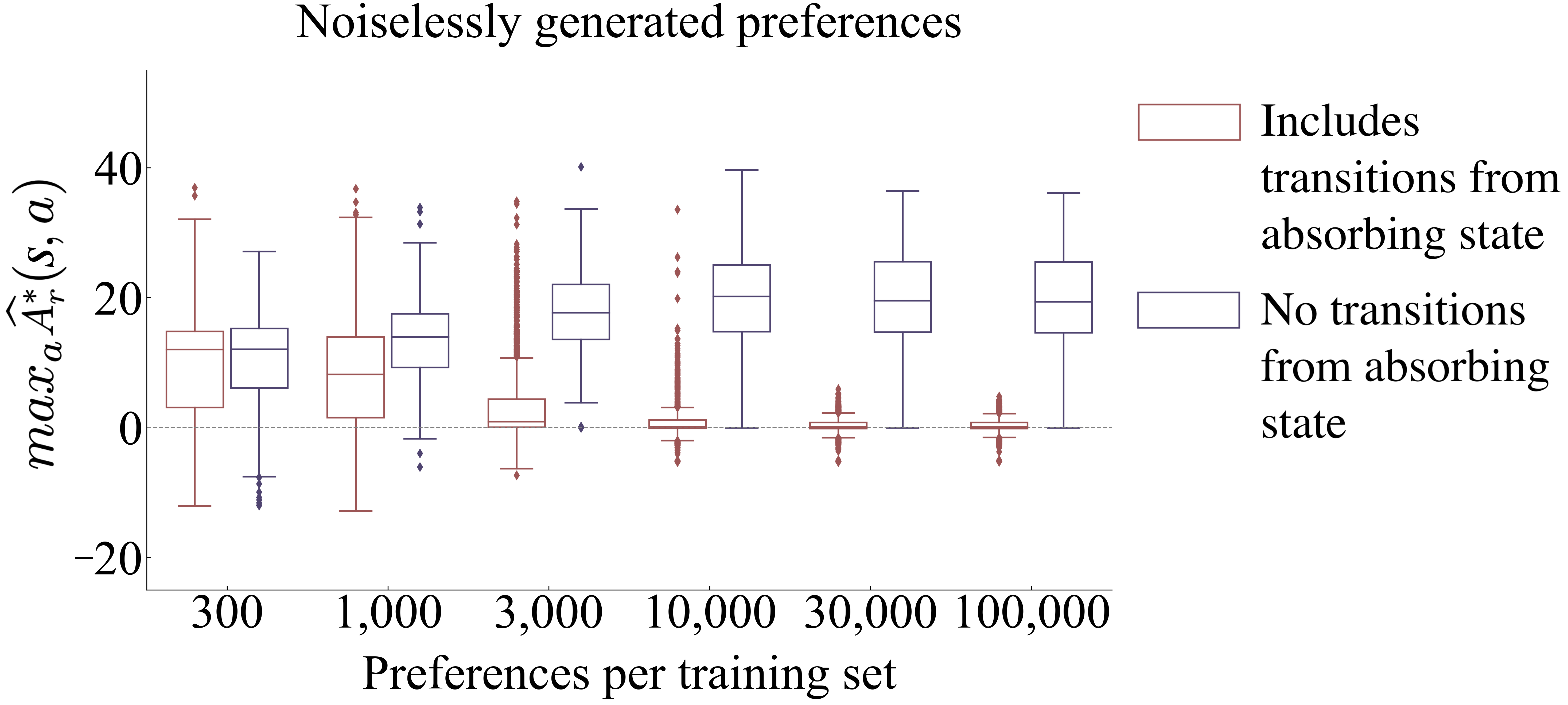}
    \vspace{-1mm}
    \caption{\footnotesize Comparing the effect on $\greedyapproxarew$ of including transitions from absorbing state. For each state within 30 MDPs, the plots above show the $max_a \approxa(s,a)$ values. The plot shows that including such transitions moves the resultant maximum values closer to 0. The plot for stochastically generated preferences is similar and can be found in Appendix~\ref{app:absorbing}. After learning with absorbing transitions, $max_a \approxa(s,a)$ across all states is stochastically closer to 0 than when learning without them. Wilcoxon paired signed-rank tests at every training set size are all extremely significant with $p < 10^{-7}$.}  
    \label{fig:absorbing_state_max_noiseless}
    \vspace{-2mm}
\end{figure}

Figure~\ref{fig:absorbing_state_max_noiseless} shows the large impact of including transitions from absorbing state when $\hat{r} = \approxa$. As expected, $\greedyapproxa$ is not noticeably affected by the inclusions of such transitions.
Further, Figure~\ref{fig:absorbing_state_max_noiseless} shows that the inclusion of these transitions from absorbing state does indeed push $max_a {A}^*_{\tilde{r}}(\cdot,a)$ towards 0, more so with larger training set sizes (given a fixed number of epochs), though it does not completely accomplish making $max_a {A}^*_{\tilde{r}}(\cdot,a)=0$.

\begin{figure}[t]
    \vspace{0mm}
    \centering
    \includegraphics[width=1.0\columnwidth]{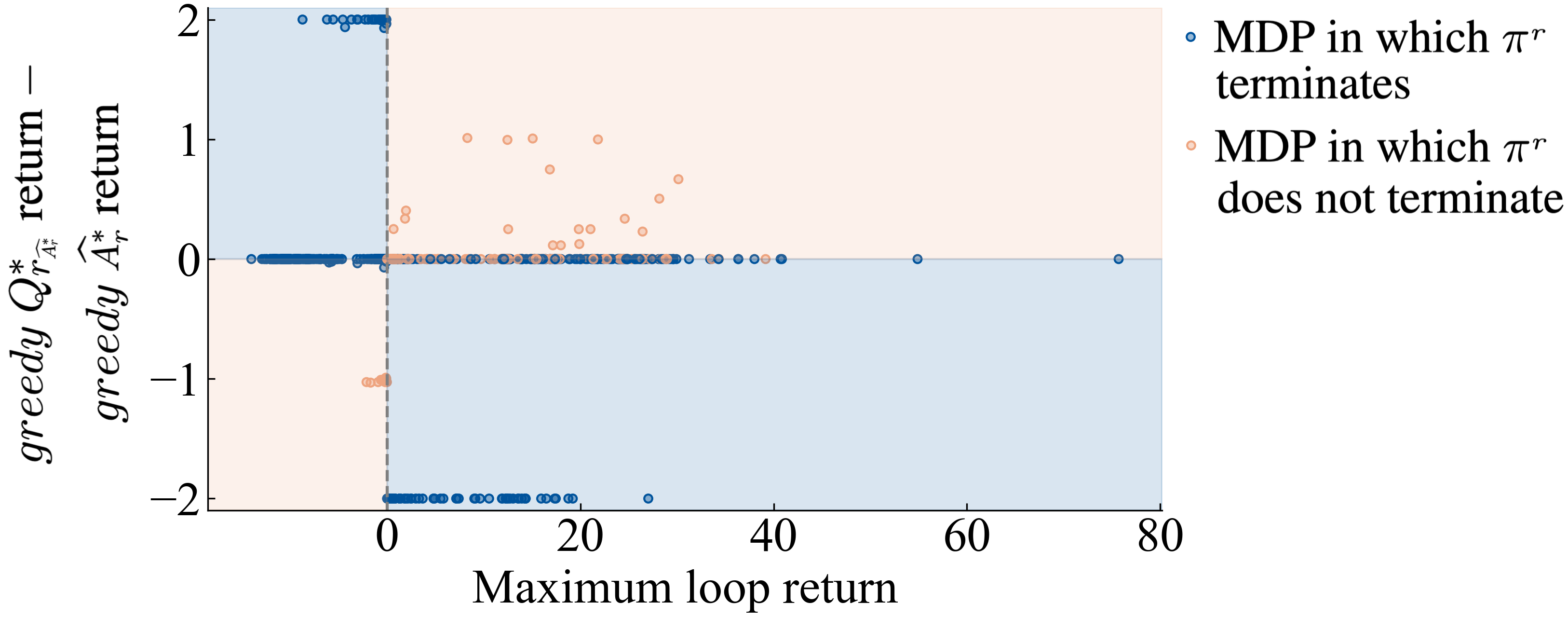}
    \vspace{0mm}
    \caption{\footnotesize 
    Validation of the hypothesis that 
    maximum partial return by $\approxarew$ across all loops determines the direction of performance differences between $\greedyapproxa$ and $\greedyapproxarew$. 1080 runs are shown,
    built from the set of 90 MDPs $\times ~\{10, 100, 1000\}$  preferences in the training set $\times ~\{1,2\}$ segment lengths $\times ~\{ \text{noiselessly},\text{stochastically}\}$ generated preferences. 
    Plot points are colored orange when every $\pi^*_r$ terminates and blue when every $\pi^*_r$ does not terminate. The blue and orange shading of the plot represents where our hypotheses predict circles of each color to be, if $y\neq0$. Returns are standardized across MDPs within $[-1,1]$ (detail in Appendix~\ref{app:experimental_settings}), and the x axis is the maximum partial return by $\approxarew$ across all loops in the MDP.
    Of the 75 runs with a performance difference ($y\neq0$), 73 conform to our hypothesis. In the remaining 2 runs, both algorithms achieve near-optimal behavior and therefore have a difference of less than 0.1.
    }
    \label{fig:max_loops}
\end{figure}

\paragraph{Bias towards termination determines  performance differences.} 
When $max_a \approxa(s,a)$ tends to be near 0, we find the performances of $\greedyapproxarew$ and $\greedyapproxa$ to be similar. 
But their performances sometimes differ. Can we predict which algorithm will perform better?
To address this questions understand why, we performed a detailed analysis with 90 small gridworld MDPs, from which the following hypothesis arose.
The logic behind the following hypothesis assumes an undiscounted task, though the hypothesized effects should exist in lessened form as discounting is increased.
We define a loop to be a segment that begins and ends in the same state and then focus on the maximum partial return by $\approxarew$ across all loops.

\begin{table}[ht]
\centering
\caption{\small Hypothesis regarding which algorithm performs as well or better than the other, given 2 conditions.}
\label{table:your_label_here}
\footnotesize{
\begin{tabularx}{\columnwidth}{|p{3.47cm}|p{1.65cm}|p{1.65cm}|}
\hline
\hspace{3.47cm} \textbf{Condition} & \textbf{\(\pi^*_r\) \mbox{terminates}} & \textbf{\(\pi^*_r\) does not \mbox{terminate}} \\
\hline
Max loop partial \mbox{return} $> 0$ & $\greedyapproxarew$ & $\greedyapproxa$ \\
\hline
Max loop partial \mbox{return} $< 0$ & $\greedyapproxa$ & $\greedyapproxarew$ \\
\hline
\end{tabularx}
}
\end{table}

Focusing on tasks with deterministic transitions,\footnote{For stochastic tasks, this concept of loops generalizes to the steady-state distribution with the maximum average reward, across all policies. 
} the justification for this hypothesis is based on the following biases created by the maximum partial return of all loops:
\begin{itemize}
    \item When the maximum partial return of all loops is \textit{positive}, any $\pi^*_{\approxarew}$ will not terminate  because it can achieve infinite value. 
    \item When the maximum partial return of all loops is \textit{negative}, any $\pi_{\hat{r}}$ for $\approxarew$ will terminate, because it can only achieve negative infinity value without terminating.
\end{itemize}

\noindent
Results shown in Figure~\ref{fig:max_loops} validate this hypothesis.
Over 1080 runs of learning $\approxa$ in various settings, 
we find that the hypothesis is highly predictive of deviations in performance.

The cause of this predictive measure, the maximum partial return by $\approxarew$ of all loops, has not yet been characterized. Hence, an algorithm designer should still be wary of mistaking $\approxa$ for a reward function and relying on this predictive measure to determine whether the resulting policy avoids or seeks termination.

\begin{figure}[t]
    \vspace{-1mm}
    \centering
    \includegraphics[width=1\columnwidth]{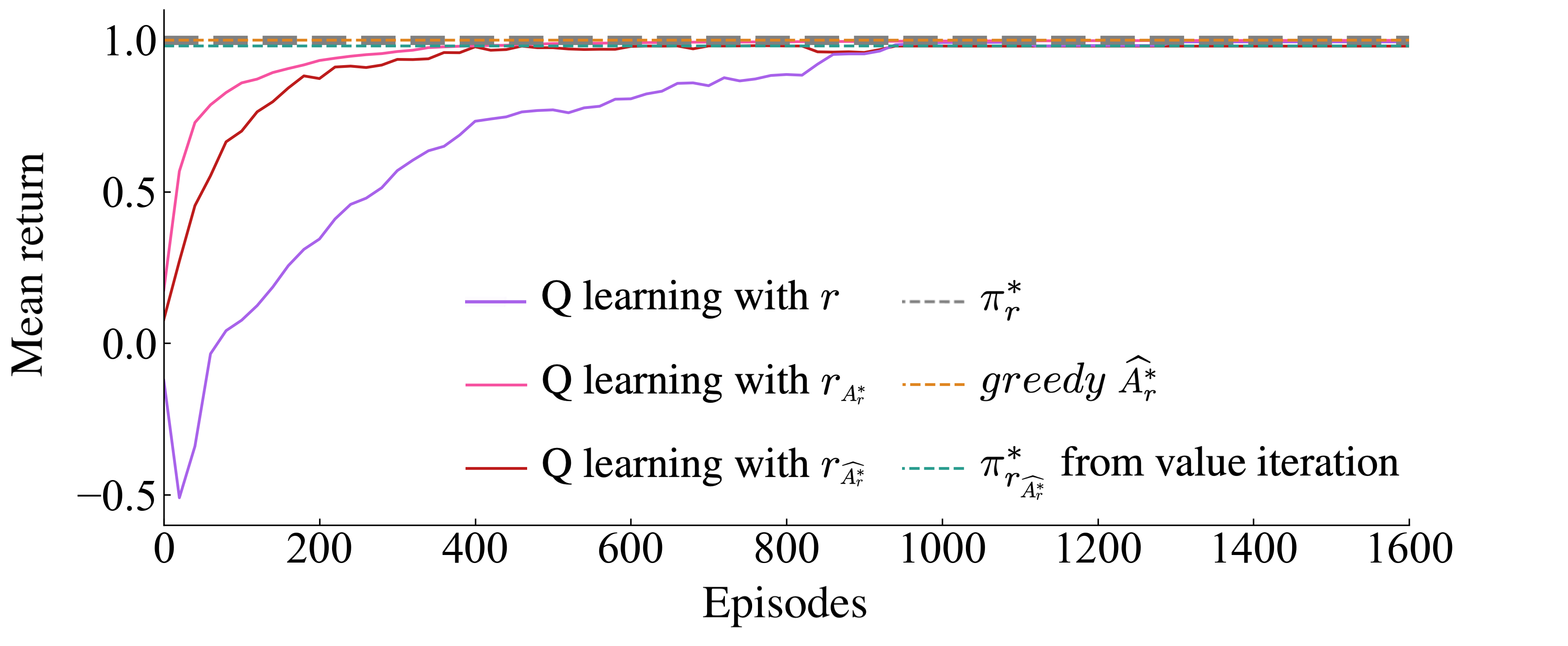}
    \vspace{-2mm}
    \caption{\footnotesize Learning curves for Q learning on the ground truth reward function $r$ and on $\approxarew$. Each curve represents 100 instances of Q learning, each in a different MDP.  $\approxa$ was learned with noiseless 100,000 regret-based preferences.
    Even without giving the learning agent access to the known $\approxarew$, we see that learning is more efficient, indicating that in practice $\approxarew$ is a helpfully shaped reward function, as is using the true $\adv$ as a reward function. We define AAC as the area above a curve and below 1.0. A small AAC indicates better learning performance. Wilcoxon paired signed-rank tests reveal that Q learning with $r$ (purple) has a larger AAC than with $\arew$ (red), which in turn has a larger AAC than with $\approxarew$ (both $p < 0.00003$).}
    \label{fig:shaping_test}
\end{figure}

\paragraph{Reward is also highly shaped with approximation error.}

We also test whether the reward shaping that exists when using $\adv$ as a reward function is also present when using its approximation, $\approxa$. Figure~\ref{fig:shaping_test} finds shows that policy improvement with the Q learning algorithm~\cite{watkins1992q} is more sample efficient with $\arew$ and with $\approxarew$ than with the ground truth $r$, as was expected.

\subsection{Summary} 
When one learns from regret-based preferences using the partial return preference model, the theoretical and empirical consequences are surprisingly less harmful than this apparent misuse suggests it would be. The policy that would have been learned with the correct regret-based preference model is preserved if $\approxa$ has a maximum of 0 in every state. Further, $\approxa$ acts as a highly shaped reward. Perhaps this analysis explains why the partial return preference model---shown to not model human preferences well~\cite{knox2022models}---nonetheless has achieved impressive performance on numerous tasks. 
That said, confusing $\approxa$ for a reward function has drawbacks compared to $\greedyapproxa$, including higher sample complexity and sensitivity to an understudied factor, the maximum partial return by $\approxarew$ of all loops.

\section{Reframing related work on fine-tuning generative models}
\label{sec:reframing}

The partial return preference model has been used in several high-profile applications: to fine-tune large language models for text summarization~\cite{ziegler2019fine}, to create InstructGPT and ChatGPT~\cite{ouyang2022training,openai2022chatgpt}, to create Sparrow~\cite{glaese2022improving}, in work by~\citet{bai2022training}, and to fine-tune Llama 2~\cite{touvron2023llama}. 
The use of the partial return model in these works fortuitously allows \textbf{an alternative interpretation of their approach: they are applying a regret preference model and are learning an optimal advantage function, not a reward function.}
These approaches make several assumptions:
\vspace{0mm}
\begin{itemize}
\item Preferences are generated by partial return.
\item During policy improvement, the sequential task is treated as a bandit task at each time step. That treatment is equivalent to setting the discount factor $\gamma$ to $0$ during policy improvement.
\item The reward function is $R \rightarrow S \times A$, not taking the next state as input.
\end{itemize}

These approaches learn $g$ as in Equation~\ref{eq:logisticonsummation}, which is interpreted as a reward function according to  the partial return preference model. They also assume $\gamma=0$ during what would be the policy improvement stage. Therefore, $\tilde{r}(s,a) = Q^*_{\tilde{r}}(s,a)$, and for any state $s$, $\pi^*_{\tilde{r}}(s) = argmax_a Q^*_{\tilde{r}}(s,a) = argmax_a \tilde{r}(s,a) = argmax_a g(s,a)$.

\paragraph{Problems with the above assumptions}

Many of the language models considered here are applied in the sequential setting of multi-turn, interactive dialog, such as ChatGPT~\cite{openai2022chatgpt}, Sparrow~\cite{glaese2022improving}, and work by \citet{bai2022training}.
Treating these as bandit tasks (i.e., setting $\gamma=0$) is an unexplained decision that contradicts how reward functions are used in sequential tasks, to accumulate throughout the task to score a trajectory as return. 

Further, the choice of $\gamma$ is arbitrary in the original framing of their algorithms. Because they also assume $|\sigma| = 1$, then the partial return of a segment reduces to the immediate reward without discounting: $\sum_{t=0}^{|\sigma|-1} \gamma^t \tilde{r}(s^{\sigma}_{t}, a^{\sigma}_{t}) = \tilde{r}(s^{\sigma}_{0}, a^{\sigma}_{0})$. Consequently, $\gamma$ curiously has no impact on what reward function is learned from the partial return preference model (assuming the standard definition in this setting that $0^0=1$). This lack of impact is a generally problematic aspect of learning reward functions with partial return preference models, since changing $\gamma$ for a fixed reward function is known to often change the set of optimal polices. (Otherwise MDPs could be solved much more easily by setting $\gamma=0$ and myopically maximizing immediate reward. \commentsig{A non-terminating MDP would be under-specified without a discount (what to maximize?) so this reads strangely as you couldn't choose a higher discount and claim to have solved the MDP. Increasing the discount from what the MDP specifies makes a solution easier to compute but it would be a solution to a different MDP and an ever worsening approximation of the solution of this MDP)}) %

Despite two assumptions---that preferences are driven only by partial return and that $\gamma=0$---that lack justification and appear to have significant consequences, the technique is remarkably effective, producing some of the most capable language models at the time of writing. 

\begin{figure}[t]
    \vspace{0mm}
    \centering
    \includegraphics[width=1.0\columnwidth]{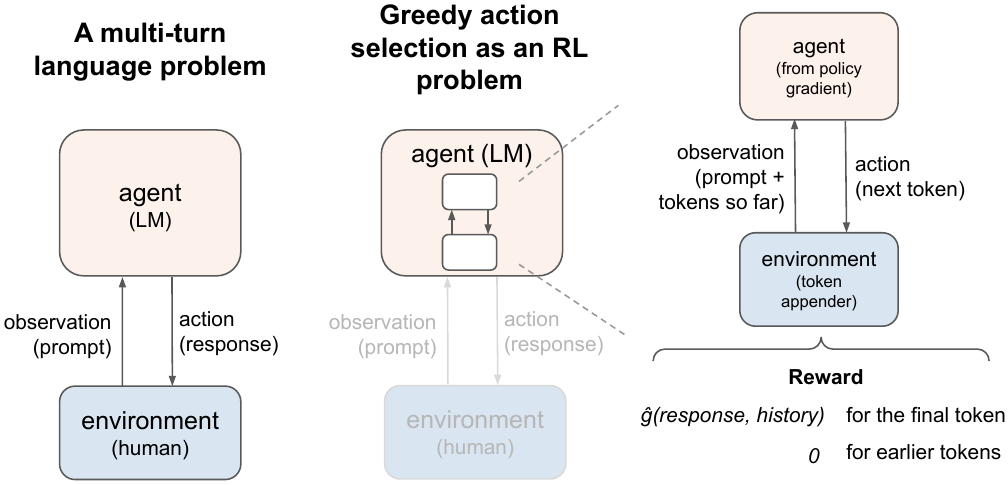}
    \vspace{0mm}
    \caption{\footnotesize This paper focuses exclusively on the problem to the left, which involves multiple turns of the human providing a prompt and the language-model agent responding. The problem on the right is a common artificial constraint on action selection to make it tractable, using the policy to sample a response one token at a time, sequentially; it does not involve any interaction with the human (i.e., the environment). %
    }
    \label{fig:2problems}
    \vspace{0mm}
\end{figure}

\vspace{-1mm}
\paragraph{Fine-tuning with regret-based preferences}
Let us instead assume preferences come from the regret preference model.
As explained in Section~\ref{sec:a-star_as_reward}, the $\gamma=0$ assumption then has no effect. Therefore it can be removed, avoiding both of the troubling assumptions. Specifically, if preferences come from the regret preference model, then the same algorithm's output $g$ is $\approxa$.  Consequently, under this regret-based framing, for any state $s$, $\pi^*_{\tilde{r}}(s) = argmax_a A^*_{\tilde{r}}(s,a) = argmax_a g(s,a)$. Therefore, \textit{both the learning algorithm and action selection for a greedy policy in this setting are functionally equivalent to their algorithm, but their interpretations change}.

In summary, assuming that learning from preferences produces an optimal advantage function---the consequence of adopting the more empirically supported regret preference model---provides a more consistent framing for these algorithms.

\vspace{-1mm}
\paragraph{A common source of confusion}
Greedy action selection can itself be challenging for large action spaces. These language models have large action spaces, since choosing a response to the latest human prompt involves selecting a large sequence of tokens. 
This choice of response is a single action that results in interaction with the environment, the human.
As an example, \citet{ouyang2022training} instead artificially restrict the selection of an action to itself be a sequential decision-making problem, forcing the tokens to be selected one at a time, in order from the start to the end of the text, as Figure~\ref{fig:2problems} illustrates. They use a policy gradient algorithm, PPO~\cite{schulman2017proximal}, to learn a policy for this sub-problem, where the RL agent receives 0 reward until the final token is chosen. At that point, under their interpretation, it receives the learned bandit reward from the left problem in Figure~\ref{fig:2problems}. This paper does not focus on \textit {how to do greedy action selection}, and we do not take a stance on whether to treat it as a token-by-token RL problem. However, if one desires to take such an approach to greedy action selection while seeking
 $\pi(s) = argmax_a [\approxa(s,a)]$, then the bandit reward is simply replaced by the optimal advantage, again executable by the same code, since both are simply the outputs of $g$.

\paragraph{Implications for fine-tuning generative models}
Extensions of the discussed fine-tuning work may seek to learn a reward function to use beyond a bandit setting. Motivations for doing so include reward functions generalizing better when transition dynamics change and allowing the language model to improve its behavior based on experienced long-term outcomes.
To learn a reward function to use in such a sequential problem setting,
framing the preferences dataset as having been generated by the regret preference model would provide a different algorithm for doing so (in Section~\ref{sec:prefmodels}). It would also avoid the arbitrariness of setting $\gamma>0$ and learning with the partial return preference model, which outputs the same reward function under these papers' assumptions regardless of the discount factor. The regret-based algorithm for learning a reward function is more internally consistent and appears to be more aligned with human stakeholder's preferences. However, it does present research challenges for learning reward functions in complex tasks such as those for which these language models are fine-tuned. In particular, the known method for learning a reward function with the regret preference model requires a differentiable approximation of the optimal advantage function for the reward function arising from parameters that change at each training iteration.

\vspace{-1mm}
\section{Conclusion}
\vspace{-1mm}
This paper investigates the consequences of assuming that preferences are generated according to partial return when they instead arise from regret. The regret preference model provides an improved account of the effective method of fine-tuning LLMs from preferences (Section~\ref{sec:reframing}). In the general case (Section~\ref{sec:oaf-as-rew}), we find that this mistaken assumption is not ruinous to performance when averaged over many instances of learning, which explains the success of many algorithms which rely on this flawed assumption. Nonetheless, this mistaken interpretation obfuscates learning from preferences, confusing practitioners' intuitions about human preferences and how to use the function learned from preferences. We believe that partial return preference model is rarely accurate for trajectory segments, i.e., it is rare for a human's preferences to be unswayed by any of a segment's end state value, start state value, or luck during transitions. Assuming that humans incorporate \textit{all} of those three segment characteristics, as the regret preference model does, results in a better descriptive model, yet it does not \textit{universally} describe human preferences. To improve the sample efficiency and alignment of agents that learn from preferences, subsequent research should focus further on potential models of human preference and also on methods for influencing people to conform to a desired preference model. Lastly, after reading this paper, one might be tempted to conclude that it's safe to close your eyes, clench your teeth, and put your faith in the partial return preference model. This conclusion is not supported by this paper, since even with the addition of transitions from absorbing states, arbitrary bias to seek or avoid termination is frequently introduced. The implication of this bias is particularly important since RLHF is currently the primary safeguarding mechanism for LLMs~\cite{casper2023open}.

\newpage
\section*{Acknowledgments}
\vspace{-2mm}

This work has taken place in part in the 
the Interactive Agents and Colloraborative Technologies (InterACT) lab at UC Berkeley, 
the Learning Agents Research Group (LARG) at UT Austin
and the Safe, Correct, and Aligned Learning and Robotics Lab (SCALAR) at The University of Massachusetts Amherst.
LARG research is supported in part by NSF (FAIN-2019844, NRT-2125858), ONR (N00014-18-2243), ARO (E2061621), Bosch, Lockheed Martin, and UT Austin's Good Systems grand challenge.
Peter Stone is financially compensated as the Executive Director of Sony AI America, the terms of which have been approved by the UT Austin. SCALAR research is supported in part by the NSF (IIS-1749204), AFOSR (FA9550-20-1-0077), and ARO (78372-CS, W911NF-19-2-0333). InterACT research is supported in part by ONR YIP and NSF HCC. Serena Booth is supported by NSF GRFP.

\bibliographystyle{plainnat}
\bibliography{references}

\newpage
\appendix
\onecolumn

\section{Proof of Theorem~\ref{thm:myopic_is_optimal}}
\label{app:theorem_proof}

\textbf{Theorem 3.1}~~(Greedy action is optimal when the maximum reward in every state is 0.)\\
\textit{$\Pi^*_{\tilde{r}} = \{\pi : \forall s, \forall a ~ [\pi(a|s) > 0 \Leftrightarrow a \in \text{argmax}_a \tilde{r}(s,a)] \}$ if $\text{max}_a \tilde{r}(\cdot,a)=0$.}

~\newline
The main idea is that if the maximum reward in every state is 0, then the best possible return from every state is 0. Therefore, $V^*_{\tilde{r}}(\cdot)=0$, making $\forall (s,a) \in S \times A, Q^*_{\tilde{r}}(s,a) = \tilde{r}(s,a) + \gamma \mathbb{E}_{s'}[V^*_{\tilde{r}}(s)] = \tilde{r}(s,a)$.

\noindent
\newline
The proof follows. 

\noindent
$\forall (s,a) \in S \times A, \tilde{r}(s,a)\leq0$, so $\forall s \in S, V^*_{\tilde{r}}(s) \leq 0$. 

\noindent
$\forall s \in S, \exists a \in A : \tilde{r}(s,a) = 0$, so $\forall s \in S, V^*_{\tilde{r}}(s) \geq 0$.

\noindent
$V^*_{\tilde{r}}(s) \leq 0$ and 
$V^*_{\tilde{r}}(s) \geq 0$ implies 
$V^*_{\tilde{r}}(s) = 0$, so $\forall s \in S, V^*_{\tilde{r}}(s) = 0$.

\noindent
$\forall (s,a) \in S \times A$,
\begin{equation}
\begin{split}
\label{eq:shift}
    Q^*_{\tilde{r}}(s,a) &= \tilde{r}(s,a) + \gamma \mathbb{E}_{s'}[V^*_{\tilde{r}}(s')] \\
    Q^*_{\tilde{r}}(s,a) &= \tilde{r}(s,a) + \gamma \mathbb{E}_{s'}[0] \\
    Q^*_{\tilde{r}}(s,a) &= \tilde{r}(s,a)\\
    argmax_a Q^*_{\tilde{r}}(s,a) &= argmax_a \tilde{r}(s,a)
\end{split}
\end{equation}

\noindent
By definition, $\Pi^*_{\tilde{r}} = \{\pi : \forall s, \pi(s) = \text{argmax}_a Q^*_{\tilde{r}}(s,a) \}$. 

\noindent Since $argmax_a Q^*_{\tilde{r}}(s,a) = argmax_a \tilde{r}(s,a)$, $\Pi^*_{\tilde{r}} = \{\pi : \forall s, \pi(s) = \text{argmax}_a \tilde{r}(s,a) \}$.
\hfill $\square$

\vspace{5mm}
\section{Proof of Corollary~\ref{thm:policy_invariance}}
\label{app:corollary_proof}

\textbf{Corollary 3.1}~~(Policy invariance of $\arew$)\\
\textit{Let $\arew \triangleq \adv$.
    If $\text{max}_a \adv(\cdot,a)=0$, 
    $\pistarsarew = \Pi^*_{r}$.}

~

\noindent
Since $\text{max}_a \adv(\cdot,a)=0$ and $\arew \triangleq \adv$, $\text{max}_a \arew(\cdot,a)=0$. 

\noindent Therefore, by Theorem~\ref{thm:myopic_is_optimal}, $\pistarsarew = \{\pi : \forall s, \forall a ~ [\pi(a|s) > 0 \Leftrightarrow a \in \text{argmax}_a \arew(s,a)] \}$.

\noindent
Also, by definition, $\Pi^*_{r} = \{\pi : \forall s, \forall a ~ [\pi(a|s) > 0 \Leftrightarrow a \in \text{argmax}_a \adv(s,a)] \}$.

\vspace{3mm}\noindent
Consequently,
\begin{equation}
\begin{split}
\label{eq:invariance}
\pistarsarew 
    &= \{\pi : \forall s, \forall a ~ [\pi(a|s) > 0 \Leftrightarrow a \in \text{argmax}_a \arew(s,a)] \} \\
    &= \{\pi : \forall s, \forall a ~ [\pi(a|s) > 0 \Leftrightarrow a \in \text{argmax}_a \adv(s,a)] \} \\
    &= \Pi^*_{r}
\end{split}
\end{equation}

\vspace{1mm}
\section{Used as reward, $\adv$ is highly shaped}
\label{app:shaped}

In Section~\ref{sec:a-star_as_reward}, we stated that following the advice below of \citet{ng1999piu} is equivalent to using $A^*_{r}$ as reward. We derive this result after reviewing their advice.

In their paper on potential-based reward shaping, the authors suggest a potent form of setting $\Phi(s)$, which is $\Phi(s) = V_{M}^*(s)$. 
Their notation includes MDPs $M$ and $M'$, 
where $M$ is the original MDP and $M'$ is the potential-shaped MDP. The notation for these two MDPs maps to our notation in that the reward function of $M$ is $r$, and we ultimately derive that  the reward function of $M'$ is $\arew$.

\citeauthor{ng1999piu}'s Corollary 2 includes the statement that, under certain conditions, for any state $s$ and action $a$, $Q_{\arew}^*(s,a) = Q_{r}^*(s,a) - \Phi(s)$.
\begin{equation}
\begin{split}
\label{eq:shaping1}
    Q_{\arew}^*(s,a) &= Q_{r}^*(s,a) - \Phi(s) \\
    Q_{\arew}^*(s,a) &= Q_{r}^*(s,a) - V_{r}^*(s) \\
    Q_{\arew}^*(s,a) &= A_{r}^*(s,a) \\
    max_a Q_{\arew}^*(s,a) &= max_a A_{r}^*(s,a) \\
    max_a Q_{\arew}^*(s,a) &= 0 \\
    V_{\arew}^*(s,a) &= 0
\end{split}
\end{equation}

Eqn~\ref{eq:shaping1} above establishes two things that will be applied within Eqn~\ref{eq:shaping2} below, that $Q_{\arew}^*(s,a) = A_{r}^*(s,a)$ and that $V_{\arew}^*(s,a) = 0$.

\begin{equation}
\begin{split}
\label{eq:shaping2}
    Q_{\arew}^*(s,a) &= \arew + \gamma \mathbb{E}_{s'}[V^*_{\tilde{r}}(s')] \\ 
    Q_{\arew}^*(s,a) &= \arew + \gamma \mathbb{E}_{s'}[0] \\ 
    Q_{\arew}^*(s,a) &= \arew \\
    A_{r}^*(s,a) &= \arew
\end{split}
\end{equation}
\hfill $\square$

\vspace{2mm}
\section{Detailed experimental settings}
\label{app:experimental_settings}

Here we provide details regarding the gridworld tasks and the learning algorithms used in our experiments. The learning algorithms described include both algorithms for learning from preferences and for policy improvement. Because much of the details below are repeated from \citet{knox2022models}, some of the description in this section is adapted from that paper with permission from the authors.

\subsection{The gridworld domain and MDP generation}
\label{app:task}

\begin{figure}[t]
    \vspace{0mm}
    \centering
    \includegraphics[width=.8\columnwidth]{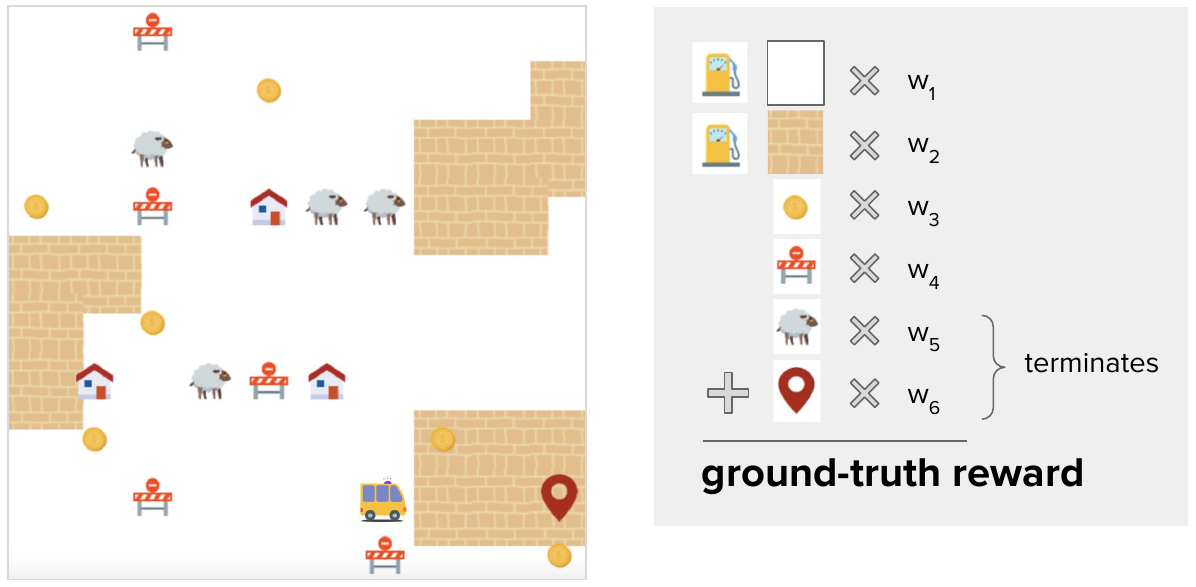}
    \vspace{0mm}
    \caption{\footnotesize An example task and unspecified reward function from the gridworld delivery domain. Tasks from this domain are used in all experiments. In Appendix~\ref{app:task}, objects described as ``mildly good'' are shown as coins, objects described as ``mildly bad'' are shown as orange road blocks, objects described as ``terminal failure states'' are shown as sheep, and objects described as ``terminal success states'' are shown as red destination markers. The brick surface and the house object were not used in our experiments. The gridworld image is reprinted with permission, from~\citet{knox2022models}.}
    \label{fig:deliver_domain}
\end{figure}

\vspace{3mm}
\paragraph{Gridworld domain}
Each instantiation of the gridworld domain consists of a grid of cells. In the following sections, each gridworld domain instantiation is referred to interchangeably as a randomly generated MDP. 

A cell can contain up to one of four types of objects: "mildly good" objects, "mildly bad" objects, terminal success objects, and terminal failure objects. Each object has a specific reward component, and a time penalty provides another reward component. The reward received upon entering a cell is the sum of all reward components. The delivery agent's state is its location. The agent's action space consists of a single step in one of the four cardinal directions. 
The episode can terminate either at a terminal success state for a non-negative reward, or at a  terminal failure state for a negative reward. The reward for a non-terminal transition is the sum of any reward components. The procedure for choosing the reward component of each cell type is described later in this subsection. 

Actions that would move the agent beyond the grid's perimeter result in no motion and receive reward that includes the current cell's time penalty reward component but not any "mildly good" or "mildly bad" components. In this work, the start state distribution is always uniformly random over non-terminal states. This domain was introduced by \cite{knox2022models}.

\vspace{3mm}
\paragraph{Standardizing return across MDPs and defining \textit{near optimal} performance}
To compare performance across different MDPs, the mean return of a policy $\pi$, $V_{r}^\pi$, is normalized to $(V_{r}^\pi - V_r^U) / V_r^*$, where $V_r^*$ is the optimal expected return and $V_r^U$ is the expected return of the uniformly random policy (both given the uniformly random start state distribution). Normalized mean return above 0 is better than $V_r^U$. Optimal policies have a normalized mean return of 1, and we consider above 0.9 to be \textit{near optimal}. 

Additionally, when plotting the mean of these standardized returns, we floor each such return at -1, which prevent the mean from being dominated by low performing policies that never terminate. Such policies can have, for example, -1000 or -10000 mean standardized return, which we group together as a similar degree of failure, yet without flooring at -1, these two failing policies would have very different effects on the means.

\vspace{3mm}
\paragraph{Generating the random MDPs used to create Figures \ref{fig:absorbing_state_noiseless}, \ref{fig:absorbing_state_max_noiseless}, and \ref{fig:shaping_test}} 
Here we describe the procedure for generating the 100 MDPs used in Figure~\ref{fig:shaping_test}, which include the 30 MDPs used in Figures \ref{fig:absorbing_state_noiseless} and \ref{fig:absorbing_state_max_noiseless}. This procedure was also used by \cite{knox2022models}.

The height for each MDP is sampled from the set $\{5, 6, 10\}$, and the width is sampled from $\{3, 6, 10, 15\}$. The proportion of cells that contain terminal failure objects is sampled from the set $\{ {0, 0.1, 0.3}\}$. There is always exactly one cell with a terminal success object. The proportion of ``mildly bad'' objects is selected from the set $\{{0, 0.1, 0.5, 0.8}\}$, and the proportion of ``mildly good'' objects is selected from $\{{0, 0.1, 0.2}\}$. Each sampled proportion is translated to a number of objects (rounding down to an integer when needed), and then each of the object types are randomly placed in empty cells until the proportions are satisfied. A cell can have zero or one object in it.

Then the ground-truth reward component for each of the above cell or object types was sampled from the following sets:

\begin{itemize}
    \item Terminal success objects: $\{{0, 1, 5, 10, 50\}}$
    \item Terminal failure objects: $\{{-5, -10, -50\}}$
    \item Mildly bad objects: $\{-2, -5, -10\}$
\end{itemize}

Mildly good objects always have a reward component of 1. An constant time penalty of -1 is also always applied.

\vspace{3mm}
\paragraph{Generating random MDPs as seen in figure \ref{fig:max_loops}}

For all 90 MDPs, the following parameters were used. The height for each MDP is sampled from the set $\{3,5\}$, and the width is sampled from $\{1,2\}$. There is always exactly one positive terminal cell that is randomly placed on one of the four corners of the board. The ground-truth reward component for the positive terminal state is sampled from $\{0,1.5,10\}$. These 90 MDPs do not contain any "mildly good" or "mildly bad" cells. 

For 30/90 of the MDPs, it is always optimal to eventually terminate at either a terminal failure cell or a terminal success cell:
\begin{itemize}
    \item For each MDP there is a $50\%$ chance that a terminal failure cell exists. If it does exist it is randomly placed on one of the four corners of the board.
    \item The ground-truth reward component for the terminal failure cell is sampled from $\{-5,-10\}$.
    \item The true reward component for blank cells is always $-1.$ 
\end{itemize}

For 30/90 of the MDPs, it is always optimal to eventually terminate at a terminal success cell:
\begin{itemize}
    \item For each MDP there is always a terminal failure cell that exists and is randomly placed on one of the four corners of the board.
    \item The ground-truth reward component for the terminal failure cell is always $-10$.
    \item The true reward component for blank cells is always $-1$. 
\end{itemize}

For 30/90 of the MDPs, it is always optimal to loop forever and never terminate:
\begin{itemize}
    \item For each MDP there is always a terminal failure cell that exists and is randomly placed on one of the four corners of the board.
    \item The ground-truth reward component for the terminal failure cell is always $-10$.
    \item The true reward component for blank cells is always $+1$. 
\end{itemize}

All parameters for randomly sampling MDPs that are not explicitly discussed above are the same as for Figures~\ref{fig:absorbing_state_noiseless}, \ref{fig:absorbing_state_max_noiseless}, and \ref{fig:shaping_test}.%

\vspace{3mm}
\subsection{Learning algorithms}

\vspace{3mm}
\paragraph{Doubling the training set by reversing preference samples} 
To provide more training data and avoid learning segment ordering effects, for all preference datasets we duplicate each preference sample, swap the corresponding segment pairs, and reverse the preference.

\vspace{2mm}
\paragraph{Discounting during value iteration and Q learning}
Despite the gridworld domain being episodic, a policy may endlessly avoid terminal states. In some MDPs, such as a subset of those used in Figure \ref{fig:max_loops}, this is an optimal behavior. In other MDPs this is the result of a low-performing policy. To avoid an infinite loop of value function updates, we apply a discount factor of $\gamma=0.999$ during value iteration, Q learning, and when assessing the mean returns of policies with respect to the ground-truth reward function, $r$. We chose this high discount factor to have negligible effect on the returns of high-performing policies (since relatively quick termination is required for high performance) while still allowing for convergence within a reasonable time.

\vspace{2mm}
\paragraph{Hyperparameters for learning  $\approxa$ as seen in Figures \ref{fig:absorbing_state_noiseless}, \ref{fig:absorbing_state_max_noiseless}, \ref{fig:max_loops}, and \ref{fig:absorbing_state_max_stochastic}} 
These hyperparameters exactly match those used in \cite{knox2022models}, except that we decreased the number of training epochs. For all experiments, each algorithm was run once with a single randomly selected seed. 
\begin{itemize}
    \item learning rate: $2$
    \item number of seeds used: $1$
    \item number of training epochs: $1,000$
    \item optimizer: Adam
    \begin{itemize}
      \item{$\beta_1=0.9$}
      \item{$\beta_2=0.999$}
      \item{eps= $1e-08$}
    \end{itemize}
\end{itemize}

\vspace{2mm}
\paragraph{Hyperparameters for learning  $\approxa$ as seen in Figure \ref{fig:shaping_test}}
These hyperparameters exactly match those used in \cite{knox2022models}. For all experiments, each algorithm was run once with a single randomly selected seed.
\begin{itemize}
    \item learning rate: $2$
    \item number of seeds used: $1$
    \item number of training epochs: $30,000$
    \item optimizer: Adam
    \begin{itemize}
      \item{$\beta_1=0.9$}
      \item{$\beta_2=0.999$}
      \item{eps= $1e-08$}
    \end{itemize}
\end{itemize}

\vspace{2mm}
\paragraph{Hyperparameters for Q learning as seen in Figure \ref{fig:shaping_test}} 
These hyperparameters were tuned on 10 learned $\approxa$ functions where setting the reward function as $\approxa$ and using value iteration to derive a policy was known to eventually lead to optimal performance. The hyperparameters were tuned so that, for each $\approxa$ function in this set, Q-learning also yielded an optimal policy. For all experiments, each algorithm was run once with a single randomly selected seed.
\begin{itemize}
    \item learning rate: $1$
    \item number of seeds used: $1$
    \item number of training episodes: $1,600$
    \item maximum episode length: $1000$ steps
    \item initial Q values: $0$
    \item exploration procedure: $\epsilon\text{-}greedy$
    \begin{itemize}
      \item{$\epsilon=0.4$}
      \item{decay=$0.99$}
    \end{itemize}
\end{itemize}

\vspace{2mm}
\paragraph{Computer specifications and software libraries used}
The compute used for all experiments had the following specification.
\begin{itemize}
    \item processor: 1x Core™ i9-9980XE (18 cores, 3.00 GHz) \& 1x WS X299 SAGE/10G | ASUS | MOBO;
    \item GPUs: 4x RTX 2080 Ti;
    \item memory: 128 GB.
\end{itemize}

Pytorch 1.7.1 \citep{NEURIPS2019_9015} was used to implement all reward learning models, and statistical analyses were performed using Scikit-learn 0.23.2 \citep{scikit-learn}.

\vspace{3mm}
\section{Shifting such that the maximum value of $\bm{{\approxa}}$ in every state is 0}
\label{app:shift}

\begin{figure}[ht]
    \vspace{0mm}
    \centering
    \includegraphics[width=.6\columnwidth]{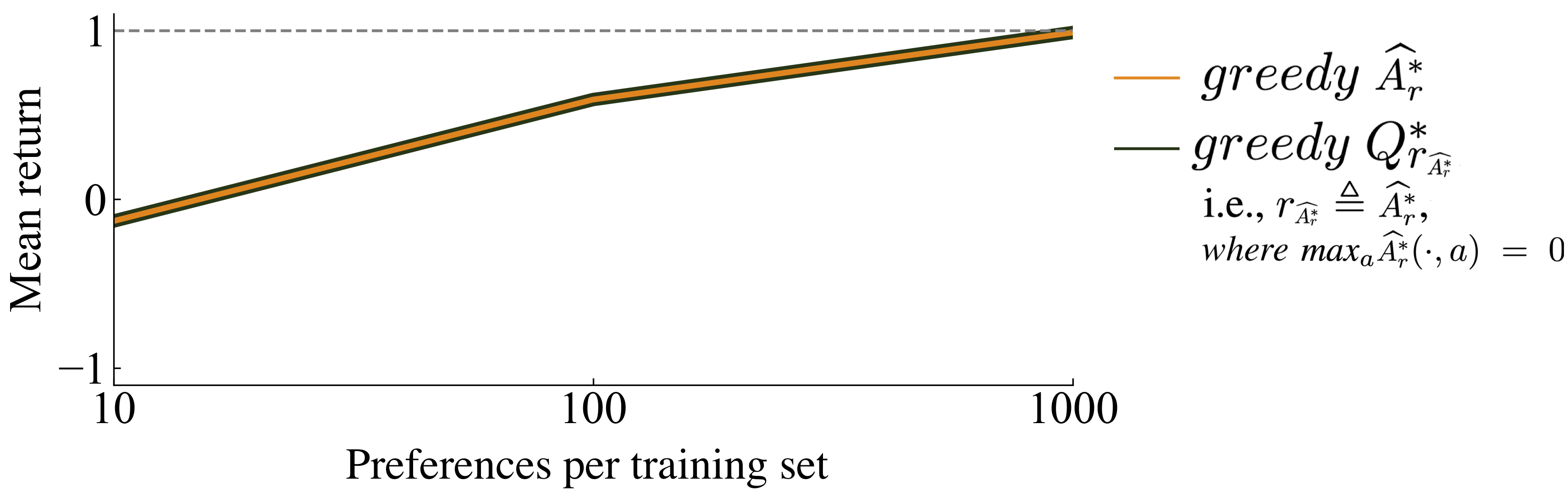}
    \vspace{0mm}
    \caption{\footnotesize In 90 small gridworld MDPs, when we explicitly shift $\approxa$ by a state-dependent constant such that $max_a \approxa(\cdot,a)=0$, we empirically observe no difference between $\greedyapproxa$ and $\greedyapproxarew$.}
\label{fig:max_a_0_confirmation}
\end{figure}

Corollary~\ref{thm:max0} claims that if $\text{max}_a \approxa(\cdot,a)=0$, then 
$\pistarsapproxarew = \{\pi : \forall s, \forall a ~ [\pi(a|s) > 0 \Leftrightarrow a \in \text{argmax}_a \approxa(s,a)] \}$. Figure~\ref{fig:max_a_0_confirmation} shows the results of our empirical validation of this claim. Specifically, this figure shows that when $\text{max}_a \approxa(\cdot,a)=0$, we observe no difference in the mean standardized return between $\greedyapproxa$ and $\greedyapproxarew$.

\vspace{3mm}
\section{Encouraging $max_a \approxa(\cdot,a)=0$ without shifting learned values manually}
\label{app:absorbing}

\begin{figure}[ht]
    \vspace{0mm}
    \centering
    \includegraphics[width=.75\columnwidth]{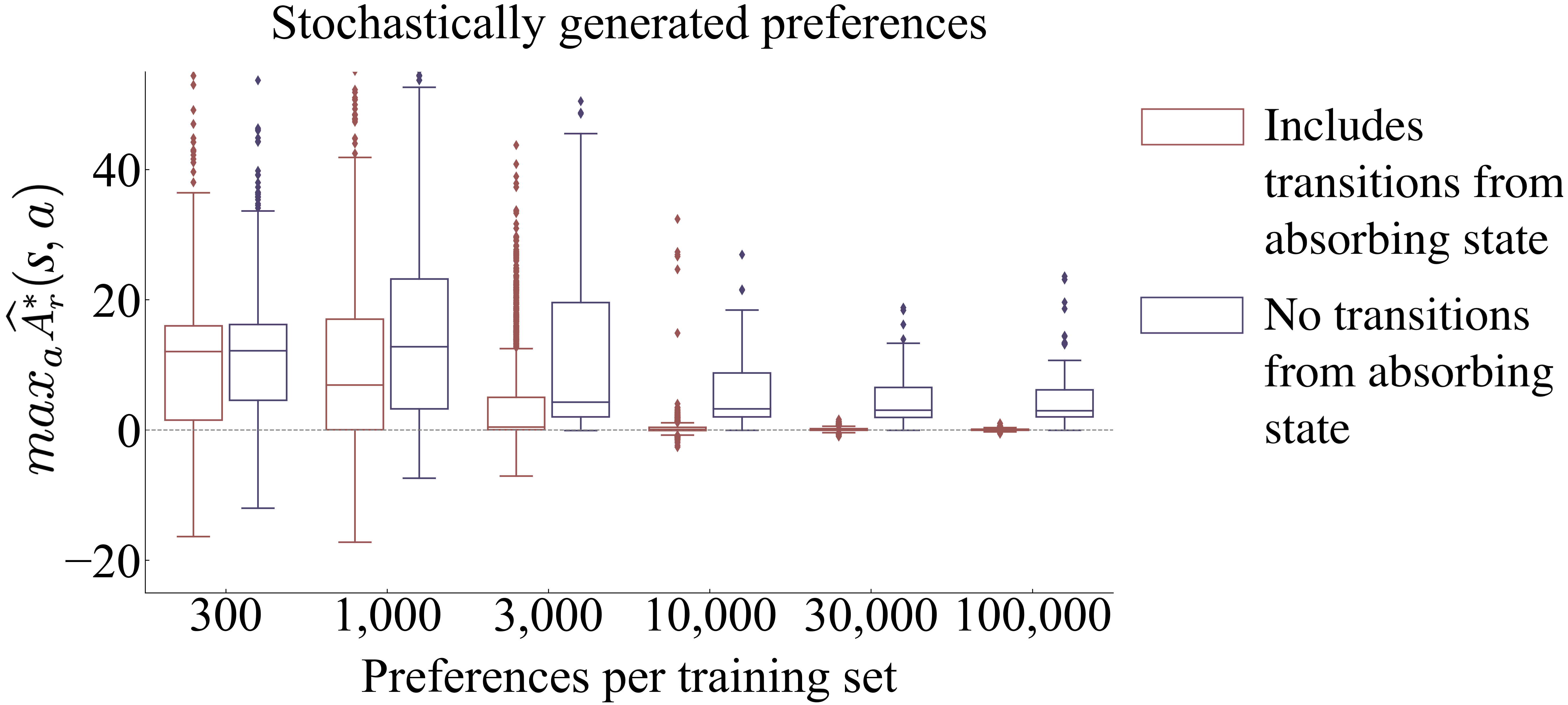}
    \\ 
    \caption{\footnotesize Comparing the effect on $\greedyapproxarew$ of including transitions from absorbing state. For each state within 30 MDPs, the plots above show the $max_a \approxa(s,a)$ values. The plot shows that including such transitions moves the resultant maximum values closer to 0. This plot complements Figure~\ref{fig:absorbing_state_max_noiseless}, which contains a similar plot for noiseless preferences. After learning with absorbing transitions, $max_a \approxa(s,a)$ across all states is stochastically closer to 0 than when learning without them. Wilcoxon paired signed-rank tests at every training set size above 300 are extremely significant with $p < 10^{-57}$. For 300 preferences, $p=0.0002$.}
\label{fig:absorbing_state_max_stochastic}
\end{figure}

Figure~\ref{fig:absorbing_state_max_noiseless} in Section~\ref{sec:approx-oaf-as-rew} uses noiselessly generated preferences. Figure~\ref{fig:absorbing_state_max_stochastic} presents an analogous analysis for stochastically generated preferences. The pattern is similar results in the noiseless setting, with even less variance for large training sets.
Specifically, Figure~\ref{fig:absorbing_state_max_stochastic} shows that including transitions from absorbing states moves the resultant maximum values of the approximated optimal advantage function closer to 0. %

\section{Investigation of performance differences}
\label{app:performance_diff_inv}

\begin{figure}[ht]
    \vspace{0mm}
    \centering    \includegraphics[width=.75\columnwidth]{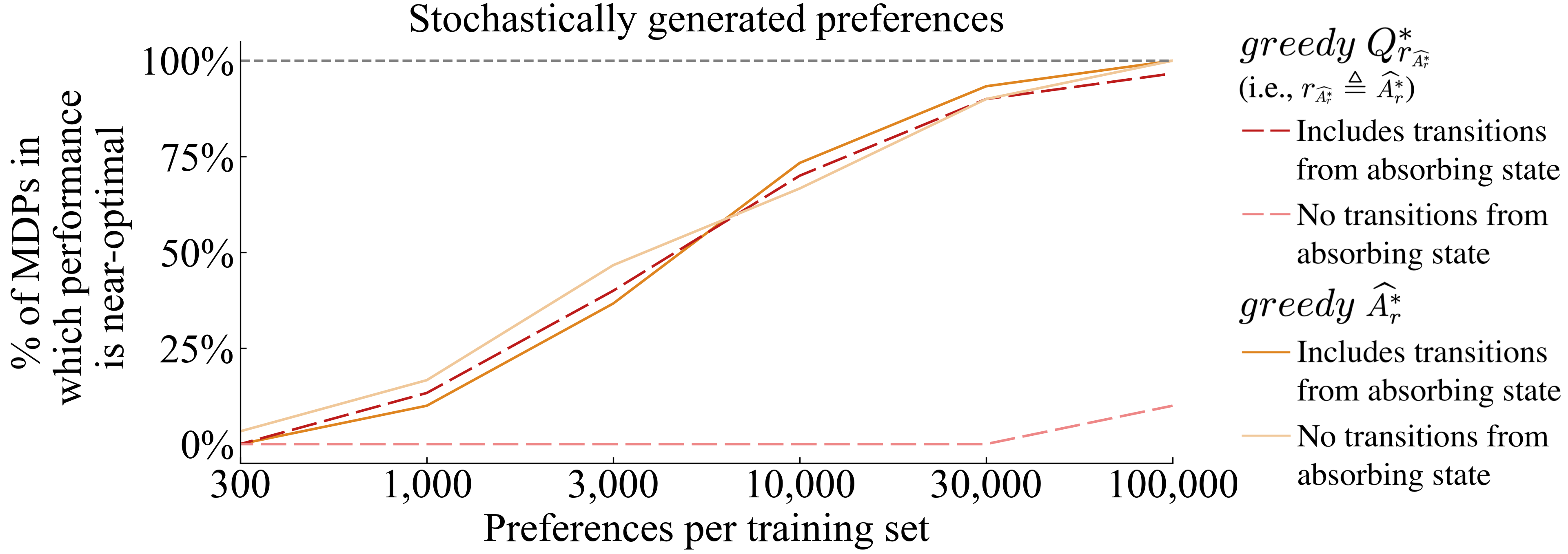}
    \vspace{0mm}
    \caption{\footnotesize Performance when stochastically generated preference datasets do and do not include segments with transitions from absorbing state, complementing Figure~\ref{fig:absorbing_state_noiseless}. Results are across 30 randomly generated gridworld MDPs with tabular representations of the $\approxa$, where segments of length 3 are chosen by uniformly randomly choosing a start state and 3 its actions.
    When transitions from absorbing states are not included, any segment that terminates before its final transition is rejected and then resampled. This plot complements Figure~\ref{fig:absorbing_state_max_noiseless}, which shows a similar plot for noiseless preferences.
    For $\greedyapproxa$ (in red) Wilcoxon paired signed-rank tests reveal that including transitions from absorbing state results in significantly higher performance for all training set sizes but the smallest, 300, with $p < 0.02$ for 1000, and $p < 0.0002$ for others. No significant difference in performance is detected for $\greedyapproxarew$ with or without terminating transitions.
    }
\label{fig:absorbing_state_stochastic}
\end{figure}

Figure~\ref{fig:absorbing_state_noiseless} in Section~\ref{sec:approx-oaf-as-rew} uses noiselessly generated preferences. As in the section above, Figure~\ref{fig:absorbing_state_stochastic} presents an analogous analysis for stochastically generated preferences. This plot likewise shows that $\greedyapproxarew$ learned without transitions from absorbing state performs poorly. We also note that the performance of the other three conditions is more similar in comparison to Figure~\ref{fig:absorbing_state_noiseless}.

\end{document}